\documentclass[lettersize,journal]{IEEEtran}
\usepackage{amsmath,amsfonts}
\usepackage{algorithmic}
\usepackage{algorithm}
\usepackage{array}
\usepackage[caption=false,font=normalsize,labelfont=sf,textfont=sf]{subfig}
\usepackage{textcomp}
\usepackage{stfloats}
\usepackage{url}
\usepackage{verbatim}
\usepackage{graphicx}
\usepackage{cite}
\usepackage{booktabs}
\begin{document}

\title{Denoised Non-Local Neural Network for \\ Semantic Segmentation}

\author{Qi Song, Jie Li, Hao Guo, Rui Huang
\thanks{This work was supported in part by Shenzhen Natural Science Foundation under Grant JCYJ20190813170601651, and in part by Shenzhen Institute of Artificial Intelligence and Robotics for Society under Grant AC01202101006 and Grant AC01202101010. \textit{(Corresponding author: Rui Huang.)}}
\thanks{Qi Song, Jie Li, Hao Guo, and Rui Huang are with School of Science and Engineering, The Chinese University of Hong Kong, Shenzhen, Guangdong, 518172, China, and also with the Shenzhen Institute of Artificial Intelligence and Robotics for Society, Guangdong, 518172, China (e-mail: qisong@link.cuhk.edu.cn; jieli1@link.cuhk.edu.cn; haoguo@link.cuhk.edu.cn; ruihuang@cuhk.edu.cn).}
}

\markboth{}%
{Shell \MakeLowercase{\textit{et al.}}: A Sample Article Using IEEEtran.cls for IEEE Journals}


\maketitle

\begin{abstract}
The non-local network has become a widely used technique for semantic segmentation, which computes an attention map to measure the relationships of each pixel pair. However, most of the current popular non-local models tend to ignore the phenomenon that the calculated attention map appears to be very noisy, containing inter-class and intra-class inconsistencies, which lowers the accuracy and reliability of the non-local methods. In this paper, we figuratively denote these inconsistencies as attention noises and explore the solutions to denoise them. Specifically, we inventively propose a Denoised Non-Local Network (Denoised NL), which consists of two primary modules, i.e., the Global Rectifying (GR) block and the Local Retention (LR) block, to eliminate the inter-class and intra-class noises respectively. First, GR adopts the class-level predictions to capture a binary map to distinguish whether the selected two pixels belong to the same category. Second, LR captures the ignored local dependencies and further uses them to rectify the unwanted hollows in the attention map. The experimental results on two challenging semantic segmentation datasets demonstrate the superior performance of our model. Without any external training data, our proposed Denoised NL can achieve the state-of-the-art performance of 83.5\% and 46.69\% mIoU on Cityscapes and ADE20K, respectively.
\end{abstract}


\section{Introduction}

\label{1}
\begin{figure}[t]
\begin{center}
\includegraphics[width=1\columnwidth]{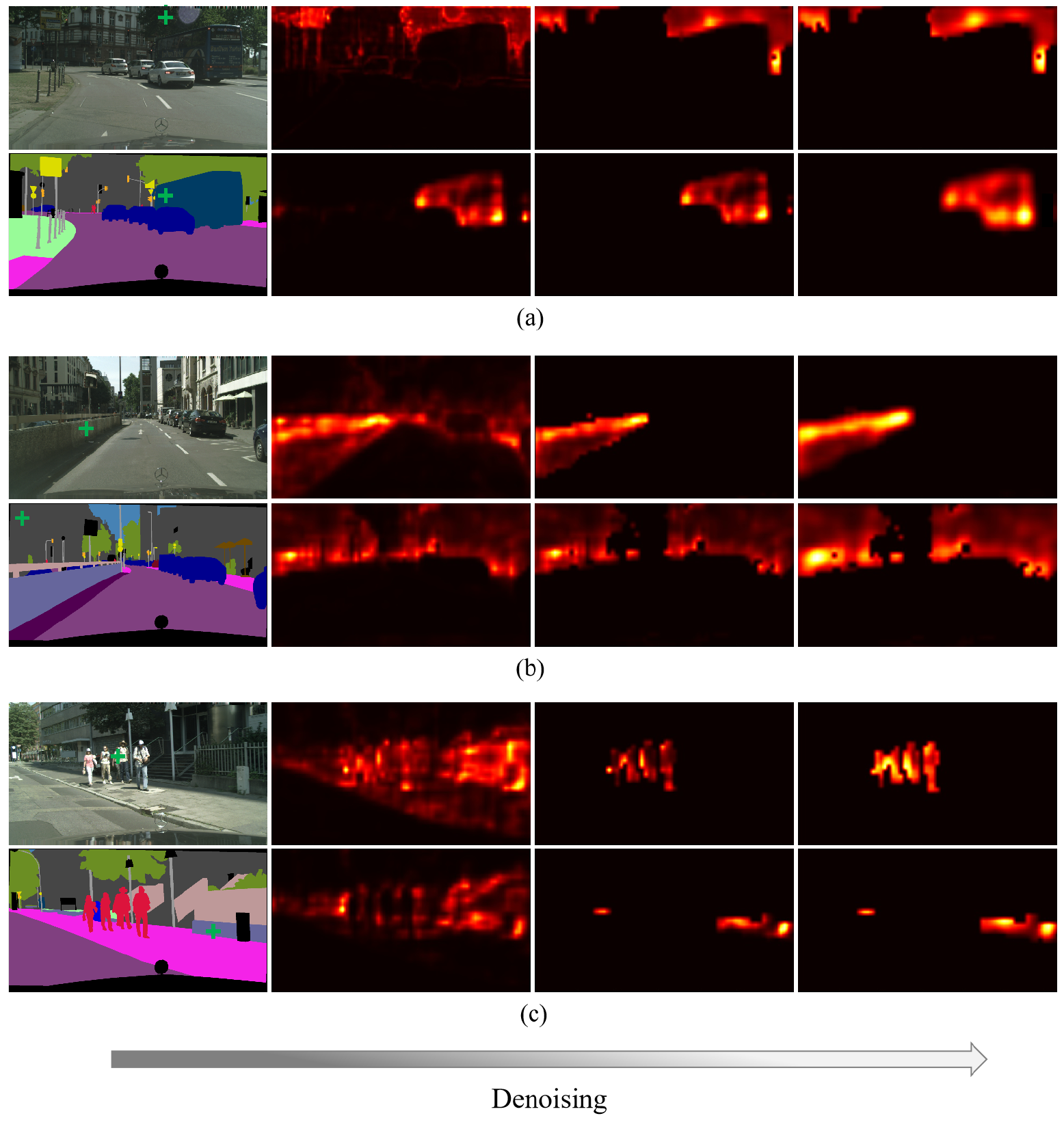} 
\end{center}
\caption{Visualization of attention maps in the non-local block and our denoised non-local block. (a)-(c) are three examples and we select two distinct pixels (marked with \textit{green cross}) for each example. The first column is the input image and its ground truth. The second column shows the original attention maps generated by the traditional non-local methods. Finally, corresponding attention maps after two-stage denoising are provided in the last two columns. }
\label{motivation}
\end{figure}

\IEEEPARstart{R}{ecently}, non-local self-attention mechanisms \cite{wang2018non, mou2019relation, fu2019dual,huang2019ccnet, zhang2019acfnet, Yuan2020ObjectContextualRF,Song2021AttaNetAN} are widely utilized in semantic segmentation to capture long-range dependencies. The typical non-local model computes the similarities between each pixel and all others via matrix multiplication and softmax operation and thus can generate an attention map of size \(N\times N\), where \(N\) is the number of pixels of the input feature. Since the input feature map always has high resolution in the semantic segmentation task, \(N\) is always a big number. It is very difficult to accurately identify the pixel-wise similarity of such a large number of pixels and different categories may bring in mixed information as well, leading to unexpected attention errors. Furthermore, we know the common knowledge that the accuracy of the calculated attention maps will directly affect the final segmentation accuracy. However, none of the previous non-local works questioned whether the calculated maps were truly reliable. As we can imagine, if the generated similarity map is untrustworthy or somewhat wrong, this will directly have the reverse effect on our final segmentation result. Based on this problem, we further investigate what the attention map captured by non-local methods actually looks like. We visualize the calculated maps and present several of them in Fig. \ref{motivation}. Theoretically, or optimally, we would like to see that an attention map of a selected point highlights only all pixels from that point's category. Nevertheless, as shown in the second column of Fig. \ref{motivation}, we can see that nearly all the maps appear to be very noisy, containing either interferences from other categories or the inconsistencies within the selected category. For example, 1) the attention map of class \textit{tree} in the first row of (a) contains interferences from class \textit{building}, 2) there exits intra-class inconsistencies of class \textit{building} in the second row of example (b). This phenomenon widely exists in the generated attention maps, and we believe it would greatly limit further performance improvements of non-local methods. In this paper, we figuratively denote these two kinds of attention flaws as attention noises, i.e., we denote the pixels that do not belong to the marked category but are highlighted as the inter-class noise and the attention flaws inside an object as intra-class noise. Based on these findings, we ask that is there any solution to denoise the coarse similarity maps? 

For the inter-class noise, we find that the final class-level predictions of the semantic segmentation network describe the overall representation of each class in an image and contain fewer inter-class noises. Therefore, we utilize the pure class-level contexts to help focus on the category information and remove the inter-class noises from the original attention maps in a global view. As for the intra-class noise, it can be seen as the hole or inconsistency that's distributed inside the object. Based on this characteristic, simply judging the category relationship between two points from a global view cannot rectify this kind of noise well. In this regard, we believe that the intra-class noise can be filled if our non-local module has a more regional/focused view. So we leverage local windows to introduce the correlations between the selected pixel and its neighbors to help our attention module further refine the details and increase the feature discriminability. To this end, we aim to denoise the pixel-wise attention map from global and local perspectives.

Our approach consists of two key components, the Global Rectifying (GR) block and the Local Retention (LR) block. In GR, we employ the class-level predictions to capture a binary map to distinguish whether the selected two pixels belong to the same category and filter the inter-class noises. In LR, we divide the contextual representations into a set of local context regions, each corresponding to the neighborhood of a specific position, and estimate the pixel-neighbor (or local) similarities to help smooth the attention hollows in the original similarity map. By applying the proposed modules, we can remove the inter-class and intra-class noises from the original attention map and generate a denoised non-local attention map that is truly reliable. As presented in the last two columns of Fig. \ref{motivation}, our model can first filter the wrong attention caused by other categories and then successfully smooth the inside inconsistencies.

\begin{figure}[t]
\begin{center}
\includegraphics[width=0.95\columnwidth]{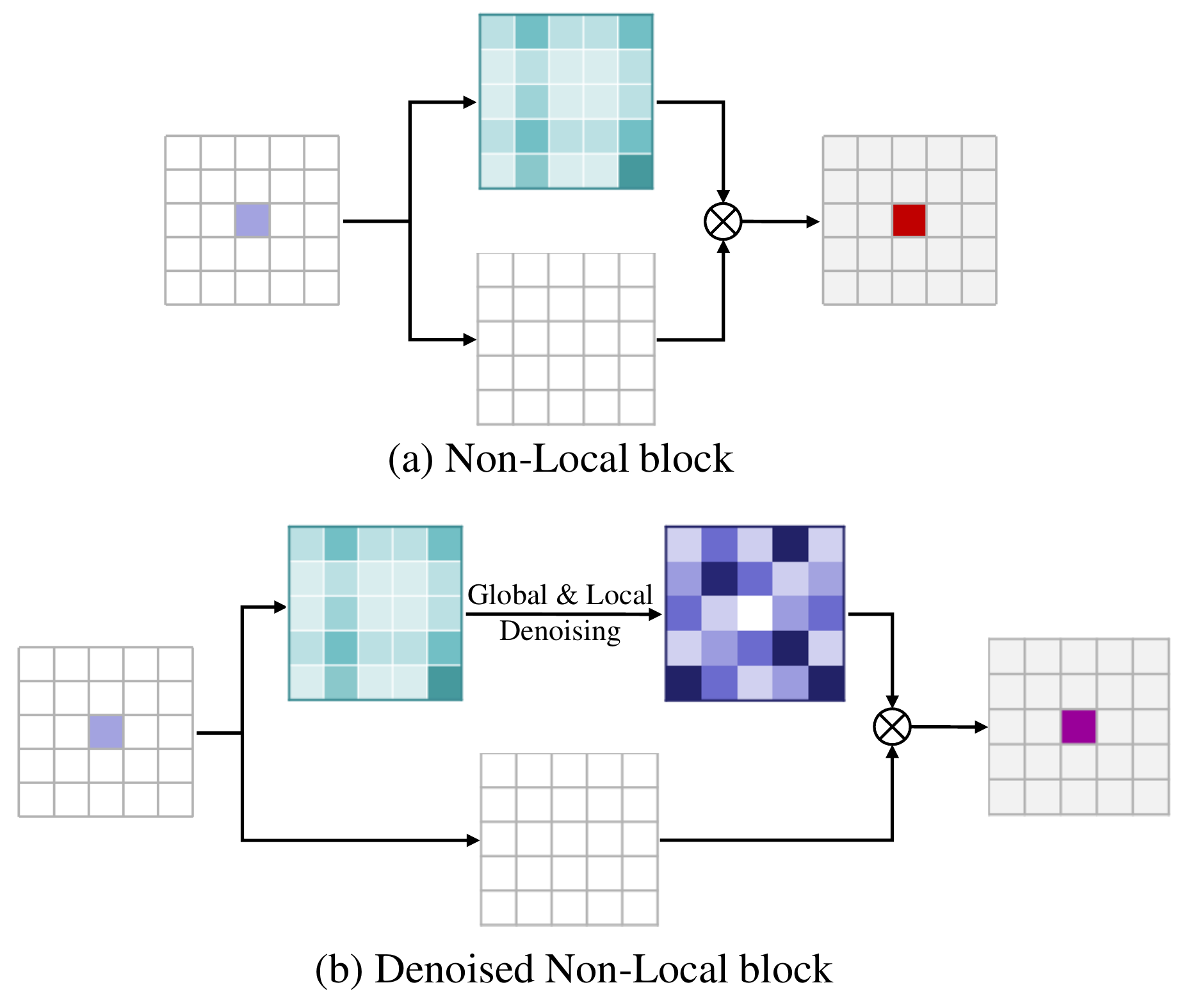} 
\end{center}
\caption{Diagrams of two non-local blocks. (a) For each position (e.g., purple), Non-local block \cite{wang2018non} generates a dense attention map (in blue), which is used to re-weight the purple position. (b) Different from the traditional non-local block, the proposed Denoised NL explores non-local mechanisms from the view of attention map denoising. Particularly, we try to remove the inter-class and intra-class noises existing in the original attention map from both global and local views. The rectified attention map is more accurate and has a more detailed view compared with the original one. Finally, the rectified similarity map is utilized to update the given pixel. For clear display, residual connections are ignored.}
\label{nlvsloe}
\end{figure}

As shown in Fig. \ref{nlvsloe}, the proposed approach differs from the traditional non-local schemes. Our model first explores the category relations of each pixel pair from the global category view to remove the inter-class noises and then leverages the predicted local similarities to further rectify the intra-class noises. However, the traditional non-local blocks measure the relationships for each pixel-pair without considering the inter-class and intra-class noises.

We have carried out extensive experiments on various challenging semantic segmentation benchmarks. Our approach outperforms the recent non-local schemes, such as DANet~\cite{fu2019dual}, CCNet \cite{huang2019ccnet}, OCNet \cite{Yuan2018OCNetOC}, and DRANet\cite{Fu2021SceneSW}. Our approach achieves top performance on the two most competitive benchmarks, i.e., 83.5\% on the Cityscapes \cite{cordts2016cityscapes} test set, and 46.69\% on the ADE20K \cite{Zhou_2017_CVPR} validation set. The main contributions of our work can be summarized as follows:
\begin{itemize}
\item We propose a novel Denoised Non-Local Network (Denoised NL) to remove the inter-class and intra-class noises of the pixel-wise similarity map. To the best of our knowledge, we are the first to explores non-local mechanisms from the view of attention map denoising.
\item The Global Rectifying (GR) block is designed to capture the class-level similarities, and the Local Retention~(LR) block is leveraged to estimate the pixel-neighbor similarities. Our method considerably enhances the segmentation results by focusing on class-specific relations and smoothing the inside hollows from global and local perspectives, respectively. 
\item The proposed method is extensively verified on two popular semantic segmentation benchmarks, including Cityscapes and ADE20K, and achieves new state-of-the-art results. 

\end{itemize}
The rest of the paper is organized as follows: Section \ref{2} presents an overview of the related works. Section \ref{3} introduces the proposed method. Section \ref{4} provides the experiments and analyses, and the conclusion is made in Section~\ref{5}.

\section{Related Work}

\label{2}
\subsection{Semantic Segmentation}
The last few years have seen a renewal of interest on semantic segmentation. Some earlier works use Conditional Random Field (CRF) \cite{Chandra2016FastEA,Chandra2017DenseAL,Chen2015SemanticIS}, Markov Random Field (MRF)~\cite{Liu2015SemanticIS} and Recurrent Neural Networks (RNNs) \cite{Liu2017LearningAV} to exploit the long-range dependencies. Different from those methods, current state-of-the-art approaches are typically based on Convolutional Neural Networks (CNNs), especially the Fully Convolution Network (FCN) frameworks \cite{long2015fully}. SegNet \cite{badrinarayanan2017segnet}, RefineNet \cite{lin2017refinenet}, UNet \cite{ronneberger2015u}, Deeplabv3+ \cite{chen2018encoder}, and DFN \cite{Yu2018LearningAD} adopt an encoder-decoder structure that fuses the multi-level information to recover the reduced spatial information. Dilated convolution \cite{Yu2016MultiScaleCA} is widely utilized to enlarge receptive fields while increasing the feature resolution. In our work, we also use dilated backbones to preserve the resolution and capture rich contextual information.

\subsection{Local and Global Context}
Context plays a critical role in various computer vision tasks, including semantic segmentation. There are plenty of works concentrating on how to take advantage of more discriminative contexts to support segmentation. Local kernels are used in convolutional layers, which encourage the network to learn local correlations. DeepLab series \cite{chen2018encoder, chen2017deeplab} propose the Atrous Spatial Pyramid Pooling (ASPP) to model the local contexts with different receptive fields. PSPNet~\cite{zhao2017pyramid} utilizes the average pooling operation to exploit global context information over four different pyramid scales. Moreover, attention mechanisms have been employed in many works~\cite{Zha2020RobustDC, Zheng2021GlobalAL, Yang2019AttentionIR} to capture multi-scale interactions. For example, SENet~\cite{hu2018squeeze} applies channel-based attention to modulate the weight of CNN channels selectively. Self-attention based models~\cite{Song2021AttaNetAN, fu2019dual,huang2019ccnet,li2019expectation} can capture pixel-wise contextual information in a non-local manner. We can also know that local contexts have a restricted receptive field (such as 3\(\times\)3) so that they cannot make the correct judgment on the far away pixels. On the other hand, global contexts enable the interactions between long-range feature points while inevitably introducing some noises or misjudgments from the isolated regions. Nevertheless, pixels are locally or globally connected in the above approaches, which cannot make full use of their complementary characteristics. In this paper, we leverage both of them to rectify the calculated attention maps rather than the feature maps. To the best of our knowledge, we are the first to utilize both local and global contexts to denoise the similarity map.

\subsection{Non-Local/Self-Attention Mechanism}
Self attention is firstly proposed in Natural Language Processing (NLP) \cite{Chorowski2015AttentionBasedMF, Cui2017AttentionoverAttentionNN,vaswani2017attention} to draw global dependencies between input and output. These years, self-attention models have been widely used in many tasks by augmenting CNNs with non-local or long-range dependencies. Non-local Network \cite{wang2018non} utilizes a self-attention mechanism to generate more powerful pixel-wise representation. DANet \cite{fu2019dual} proposes two types of attention modules to model both spatial and channel dependencies. CCNet \cite{huang2019ccnet} harvests long-range relations of its surrounding pixels on the criss-cross path through a novel crisscross attention module. ACFNet \cite{zhang2019acfnet} explores class-level context with the usage of self-attention mechanism. These methods model the relationships between pixel pairs with non-local receptive fields. Recently, work \cite{Ramachandran2019StandAloneSI} propose a stand-alone self-attention method to replace the traditional convolution operation and build a fully self-attentional model. However, we can find that previous approaches all assumed that the generated attention map is reliable and did not further investigate the information captured by the similarity map. In this work, we question the reliability of the computed attention map and improve the traditional non-local mechanism from the view of attention map denoising. 

\begin{figure*}[t]
\begin{center}
\includegraphics[width=0.95\textwidth]{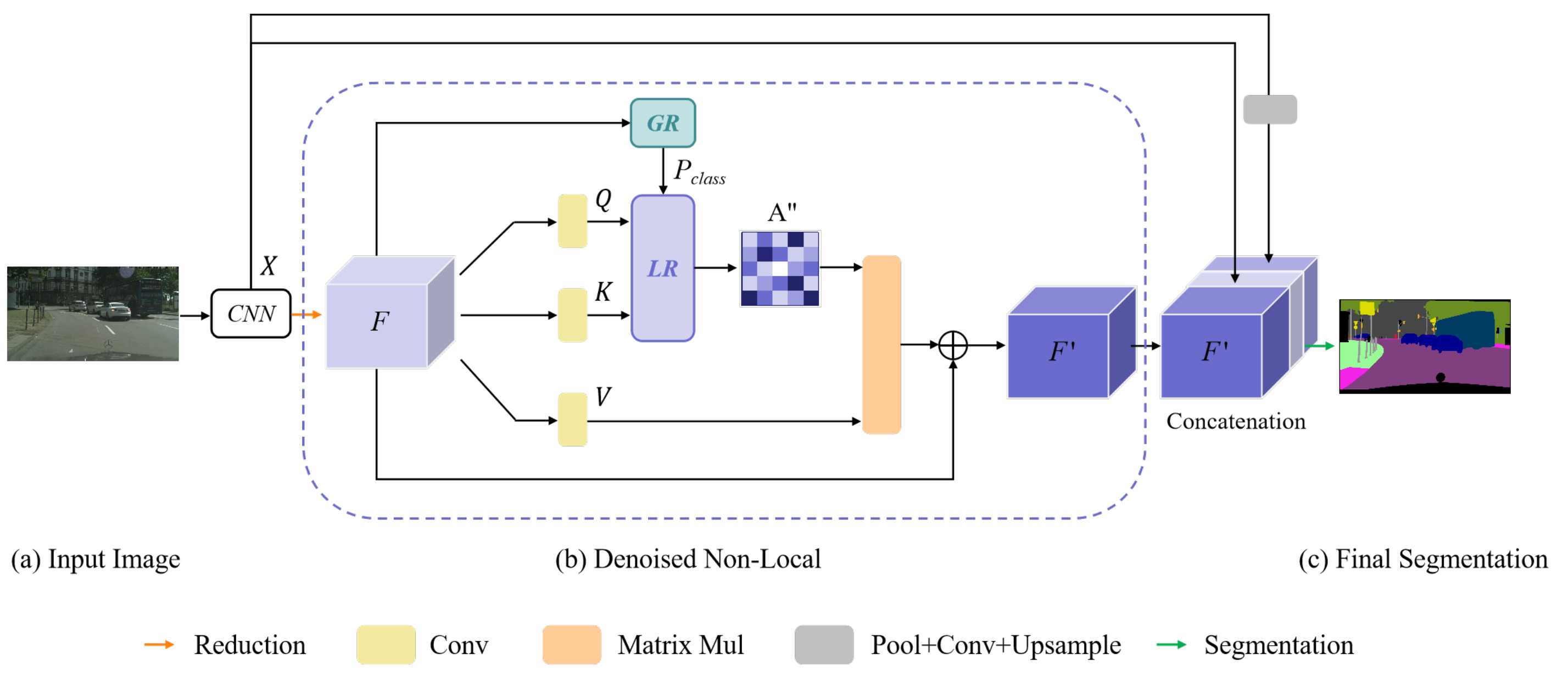} 
\end{center}
\caption{Illustration of the overall architecture. Two major components are involved, i.e., Global Rectifying (GR) block and Local Retention (LR) block, which are used to remove both inter-class and intra-class noises of the original non-local attention map.}
\label{framework}
\end{figure*}

\section{Methodology}

\label{3}
In this section, we first describe some background of our method in the next subsection. Then the entire pipeline of our proposed framework is introduced in Sec. \ref{3.2}. After that, we will give the detailed description of each component in Sec.~\ref{3.3}-\ref{3.4}. Finally, the network details are presented in Sec.~\ref{3.5}.

\subsection{Background}
\textbf{Context Modeling.} Both traditional convolutions and dilated convolutions are employed with multi-scale neighborhoods (i.e., kernel sizes \(k\) with or without dilation) to model local relations. Given an input  \(x\in \mathbb{R}^{C_{in} \times H\times W}\)  with the learned weight matrix \(W^d\in \mathbb{R}^{C_{out} \times C_{in} \times k\times k}\) , where \(C_{in}\) and \(C_{out}\) are the number of input and output channels, \(H\) and \(W\) are the spatial dimensions of the input tensor. The output \(y_{ij}\in \mathbb{R}^{C_{out}}\) at position \(ij\) is defined by:
\[y_{ij}^d = \sum W^d x_{ab}\,,\eqno{(1)}\]

Here, \(x_{ab}\) is the sampled pixel for the dilation convolution with the dilation rate \(d\). \(y_{ij}^d\) is the corresponding output representation at position \(ij\). Although the dilated convolutions can enlarge the receptive fields and capture multi-scale contextual information, they still lack dense and long-range correlations with other isolated pixels.

\textbf{Self-Attention.} Self attention is widely used in vision tasks to augment convolutional representations. Given an input feature map \(x\in \mathbb{R}^{C_{in} \times H\times W}\), the output at position \(ij\) is computed by:
\[y_{ij} = \sum softmax_{ab}(Q^T_{ij}K_{ab})V_{ab}\,,\eqno{(2)}\]

where the queries \(Q\), keys \(K\), and values \(V\) are linear projections of the input feature. The \(softmax\) denotes a softmax function applied to all possible positions, from which a pixel-wise attention map is calculated. 

The above self-attention mechanism updates the original features via aggregating features at all positions with the weighted summation, where the weights are calculated by the feature similarities between the corresponding two positions. This scheme models the pixel-wise dependencies with non-local receptive fields and the generated similarity map is used to augment convolutional features. In this way, non-local relationships are encoded into the whole feature map instead of convolution, which only captures local relations.

Recently, work \cite{Ramachandran2019StandAloneSI} proposes a stand-alone self-attention mechanism to replace the spatial convolutions. Given a pixel  \(x_{ij}\in \mathbb{R}^{C_{in}}\), they first extract a local region with spatial kernel \(k\) centered around \(x_{ij}\). Then, the relations between the pixel and its neighbors are generated via the softmax operation, which are further used as the weight maps. Like regular convolutions, the output is the inner product between the local window and the learned weights, and the fully self-attentional layer still sacrifices the global dependencies with other isolated regions.

\subsection{Overall Architecture}
\label{3.2}
From the above analysis, we can find that previous self-attention methods augment convolutional features with independent local or global dependencies without the consideration of noises contained in the similarity map. Different from the previous self-attention methods, we propose a novel Denoised Non-Local (Denoised NL) to improve the traditional non-local mechanism from the view of attention map denoising. To the best of our knowledge, we are the first to explore non-local mechanisms towards this research direction.

The network architecture is given in Fig. \ref{framework}. An input image is passed through a Convolutional Neural Network (CNN) and produces a feature map \(X\). After obtaining the feature map \(X\), we first feed \(X\) into a convolution layer and a Basic Block to reduce the channel dimension and the spatial size (see Reduction in Fig. \ref{framework}), resulting in a feature map \(F\). In the Basic Block, a \(3\times 3\) convolutional layer with stride 2 is first used to reduce the spatial size, then a \(3\times 3\) convolutional layers is used to enrich semantic information, and the residual connection is used to alleviate the gradient vanishing/exploding problem. Our proposed Denoised NL is then applied on the feature \(F\), which can capture more powerful dependencies without the interference of attention noises. Therefore, each position in the output feature \(F'\) is augmented with more reliable similarities and can successfully perceive accurate responses of all the other positions.

Finally, we concatenate \(X\) with the augmented feature map~\(F'\) generated by Denoised NL and the global context obtained by applying the global average pooling followed with a~\(1\times 1\) convolution on \(X\) to make the final prediction. Semantic predictions are performed by a convolutional head consisting of (\(3\times 3\) Conv)\(\to\)(BN)\(\to\)(ReLU)\(\to\)(\(1\times 1\) Conv), as shown in Fig. \ref{framework} Segmentation. 

\subsection{Denoised NL}

The detailed architecture of Denoised NL can be seen in the middle part of Fig. \ref{framework}. Our approach contains two main components, i.e., Global Rectifying (GR) block and Local Retention (LR) block. GR employs the class-level predictions to capture a binary map, which is applied to distinguish whether the selected two pixels belong to the same category and remove the inter-class noises. Furthermore, LR structures the contextual pixels into local context regions and calculates the local attention map to rectify the holed or inconsistent attention mistakes inside objects (i.e., intra-class noises). In this way, our module has a more well-rounded and detailed contextual view compared to previous methods and can achieve mutual gains for semantic segmentation. 

Given an input feature map \(F\in \mathbb{R}^{C\times H \times W}\), where \(C\) is the number of channels, \(H\) and \(W\) are the spatial dimensions of the input tensor. We first feed \(F\) into the GR block to calculate the binary map \(P_{class}\), which will be further used to remove the inter-class noises in the similarity map. Meanwhile, we feed \(F\) into three convolution layers with \(1\times1\) filters to generate three new feature maps \(Q\), \(K\), and \(V\) respectively, where \(\{Q, K\}\in \mathbb{R}^{C'\times H \times W}\) and \(V\in \mathbb{R}^{C\times H \times W}\). \(C'\) is less than \(C\) due to dimension reduction. We then reshape feature maps \(Q\), \(K\), and \(V\) to \(\mathbb{R}^{N\times C'}\), \(\mathbb{R}^{C'\times N}\), and \(\mathbb{R}^{N\times C}\) respectively, where \(N = H\times W\) is the number of all pixels. Next, we send feature maps \(Q\) and \(K\) to the LR block to fill the attention hollows inside objects with the usage of pixel-neighbor relations. After the denoising operation of GR and LR, the newly generated attention map \(A''\) is more compact and accurate than the original one, which is further applied to update feature weights of \(V\) via matrix multiplication. Finally, we perform an element-wise sum operation with the input features \(F\) to obtain the final output \(F'\in \mathbb{R}^{C\times H\times W}\) as follows:
\[F'_j = \gamma \sum_{i=1}^N(a''_{ji}V_i)+F_j\,\eqno{(3)}\]

in which \(a''_{ij}\in A''\) denotes the degree of correlation between position \(i\) and \(j\). \(\gamma\) is the learned weights to balance the augmented features and original features. Equation (3) shows that the pixel-wise dependencies between different positions after denoising are used to augment the input feature.

\begin{figure}[t]
\begin{center}
\includegraphics[width=1\columnwidth]{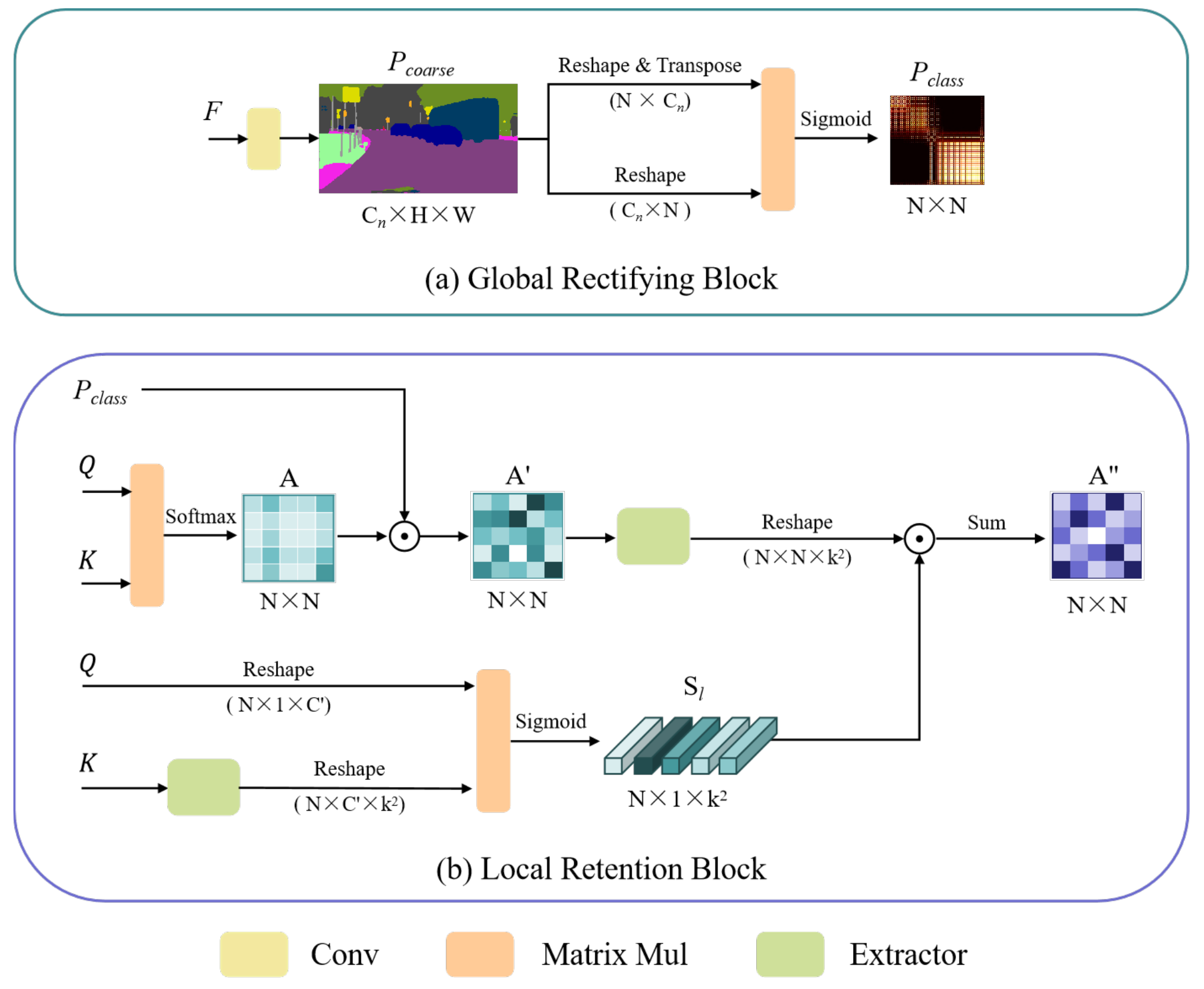} 
\end{center}
\caption{Details of Global Rectifying block and Local Retention block. Extractor represents the operation to extract local sliding windows.}
\label{DNL}
\end{figure}

\subsection{Global Rectifying Block}
\label{3.3}
In Section \ref{1}, we find the phenomenon that some pixels that do not belong to the marked category are always highlighted in the attention map, which is a kind of attention mistake and is denoted as the inter-class noise in our paper. To remove the inter-class noises, we design the Global Rectifying block and present its architecture in Fig. \ref{DNL} (a). Particularly, we find that the class-level predictions of the semantic segmentation network describe the representation of each class and contain fewer inter-class noises. Therefore, we leverage the pure class-level contexts to help focus on the category information and remove the inter-class noise from the generated attention maps in a global view. Given an input feature \(F\in \mathbb{R}^{C\times H \times W}\), we first predict the coarse segmentation result \(P_{coarse} \in \mathbb{R}^{C_n\times H \times W}\) via a \(1\times 1\) convolutional layer, where \(C_n\) is the number of the classes.

To further decide that whether the selected pixel pair belong to the same class or not, we reshape the \(P_{coarse}\)  to \(\mathbb{R}^{C_n\times N}\) and conduct the matrix multiplication between \(P_{coarse}\) and its transpose. Then we employ a sigmoid operation on the result, leading to the class-specific binary map \(P_{class} \in \mathbb{R}^{N\times N}\). When calculating \(P_{class}\), we choose to use the sigmoid function rather than the softmax operation because we only want to figure out the categorical relation between pixel pairs and maximize the difference between different categories.

\subsection{Local Retention Block}
\label{3.4}
When analyzing the intra-class noise, we find that it is generally displayed as the holes in the attended regions inside the object. In this situation, the attention weight of this holed region tends to be 0 ($\rightarrow$0). Though GR can correctly judge that the noisy position is with high attention weight, simply multiplying it cannot recover the weight of the holed region. In a word, multiplying holes with attention weights generated from GR cannot sufficiently solve the intra-class noise.

Besides, we observe that a local region often contains detailed textures and spatial relations, while the traditional self-attention methods ignore this characteristic introduced by the 2D relation position and focus on the pixel-wise long-range similarities. We believe the local information can help correct the misclassified feature weights inside objects. Motivated by this, we propose the Local Retention block to fill the attention hollow with the dependencies between each pixel and its local region. Fig. \ref{DNL}~(b) gives the detailed settings of our Local Retention block.

Given three input features, we first perform a matrix multiplication between \(Q\) and \(K\) followed by a softmax layer to learn original similarities \(A\in \mathbb{R}^{N\times N}\) between all pixel pairs:
\[ A_{ij}=\frac{exp(Q_{i}\cdot K_{j})}{\sum_{i=1}^N exp(Q_{i}\cdot K_{j})}\,,\eqno{(4)}\]

where \(A_{ij}\) measures the \(i^{th}\) pixel's impact on the \(j^{th}\) pixel and \(A\) is a defective attention map with both inter-class and intra-class misjudgement. 

After that, we apply an element-wise multiplication between \(A\) and the class-specific binary map \(P_{class}\) generated from GR block to eliminate the inter-class noises, yielding a new attention map \(A'\). Meanwhile, to capture the pixel-neighbor (local) similarity, we utilize the extractor operation similar to the sliding window in convolutions to extract local neighborhoods (i.e., \(k\times k\)) around each pixel of feature \(K\). In this way, we can restrict the receptive field of self-attention to a local square region and let our model learn to rectify inside attention hollows with the help of neighboring information. We further reshape \(Q\) and \(K\) to \(\mathbb{R}^{N\times 1\times C'}\) and \(\mathbb{R}^{N\times C'\times k^2}\), where \(k\) is the kernel size of local region. Then, we conduct the matrix multiplication followed by a sigmoid function between \(Q\) and \(K\) to calculate the relationships between pixels and their local neighbors, i.e., \(S_l\).

In the LR block, we view the local region within a kernel window as a whole and leverage the interaction inside a local region to help correct the inner attention mistakes. Specifically, we extract the local regions of \(A'\) and then multiply it with the local similarities \(S_l\) to update each attention weight with a weighted sum of its local dependencies, leading to a compact attention map \(A''\) without attention noises. 

\subsection{Network Details}
\label{3.5}
We use ResNet-101 \cite{he2016identity} pre-trained from ImageNet \cite{russakovsky2015imagenet} as the backbone network. Noted that we remove the down-sampling operations and employ dilated convolutions in the last two ResNet blocks, thus enlarging the size of the final feature map size to 1/8 of the input image. However, the input resolution of our Denoised NL is 1/16 of the original image by applying the down-sampling operation, i.e., the Reduction in Fig. \ref{framework}.

For explicit feature refinement, we use deep supervision to get better performance and make the network easier to optimize. The principal loss function is applied to supervise the output of the whole network. Moreover, we also use a linear function to predict the segmentation results from the output feature of the res2 block and res3 block supervised with pixel-wise cross-entropy losses, which are used as the auxiliary loss functions of our model. Finally, two parameters \(\lambda_1\) and \(\lambda_2\) are utilized to balance the principal loss and the auxiliary loss:
\[\mathcal{L}=\mathcal{L}_p+\lambda_1 \mathcal{L}_1 + \lambda_2 \mathcal{L}_2\,,\eqno{(5)}\]

where \(\mathcal{L}\) is the joint loss. \(\mathcal{L}_p\), \(\mathcal{L}_1\), and \(\mathcal{L}_2\)  are the principal loss of the final output, auxiliary loss for the output of the res2 block and res3 block, respectively. Particularly, all the loss functions are cross-entropy losses. We empirically set the weights \(\lambda_1\) and \(\lambda_2\) as: \(\lambda_1=\lambda_2=1.0\).

\section{Experiments}

\label{4}
In this section, we first introduce implementation details. Then we carry out a series of comparison and ablation experiments to verify the superiority of the proposed method on the Cityscapes \cite{cordts2016cityscapes}. Finally, we report our results on the ADE20K \cite{Zhou_2017_CVPR} dataset.

\begin{table*}[htp]
\caption{Per-class results on Cityscapes test set with the state-of-the-art models. Denoised NL outperforms existing approaches and achieves 83.5\% in Mean IoU without using extra coarse annotated data.}
\begin{center}
\resizebox{\textwidth}{!}{
\begin{tabular}{l|ccccccccccccccccccc|c}
\toprule[2pt]
Method &  road &  swalk &  build  & wall  & fence &  pole  & tlight  & sign &  veg.  & terrain  & sky  & person  & rider  & car &  truck  & bus &  train &  mbike &  bike &  mIoU (\%)\\
\midrule[2pt]
RefineNet \cite{lin2017refinenet} &98.2 & 83.3& 91.3 &47.8 &50.4 &56.1  &66.9 &71.3 &92.3 &70.3 & 94.8 &80.9& 63.3 &94.5 &64.6 &76.1 &64.3& 62.2& 70.0 &  73.6 \\
DUC \cite{Wang2018UnderstandingCF} &98.5 & 85.5& 92.8  &58.6 &55.5 &65.0 &73.5 &77.9 &93.3 &72.0 & 95.2 &84.8& 68.5 &95.4 &70.9 &78.8 &68.7& 65.9& 73.8 &  77.6 \\
ResNet-38 \cite{wu2019wider} & 98.5& 85.7& 93.1& 55.5 &59.1& 67.1& 74.8& 78.7& 93.7& 72.6& 95.5& 86.6& 69.2& 95.7& 64.5& 78.8& 74.1& 69.0& 76.7& 78.4\\
PSPNet  \cite{zhao2017pyramid}&98.6 & 86.2& 92.9 &50.8 &58.8& 64.0 & 75.6 &79.0& 93.4& 72.3 &95.4& 86.5& 71.3 &95.9 &68.2 &  79.5  &73.8 &69.5 & 77.2&  78.4 \\
SegModel \cite{shen2017semantic} & 98.6& 86.4& 92.8& 52.4& 59.7& 59.6& 72.5& 78.3& 93.3& 72.8& 95.5& 85.4& 70.1& 95.6& 75.4& 84.1& 75.1& 68.7& 75.0& 78.5\\
BiSeNet \cite{yu2018bisenet}& -& -& -& -& -& -& -& -& -& -& -& -& -& -& -& -& -& -& -&  78.9\\
AAF \cite{Ke_2018_ECCV}&98.5& 85.6 & 93.0& 53.8 &58.9 &65.9& 75.0 &78.4 & 93.7 &72.4  &95.6 &86.4 &70.5  &95.9& 73.9 & 82.7 & 76.9  & 68.7 &76.4&  79.1 \\
DFN \cite{Yu2018LearningAD}& -& -& -& -& -& -& -& -& -& -& -& -& -& -& -& -& -& -& -& 79.3\\
PSANet \cite{zhao2018psanet}& -& -& -& -& -& -& -& -& -& -& -& -& -& -& -& -& -& -& -&  80.1\\
AttaNet\cite{Song2021AttaNetAN}&98.7& 87.0& 93.5&55.9&62.6& 70.2& 78.4& 81.4& 93.9& 72.8& 95.4& 87.9& 74.7& 96.3& 71.2& 84.4& 78.0& 68.6& 78.2&  80.5\\
DenseASPP \cite{Yang2018DenseASPPFS} &98.7 &87.1& 93.4& 60.7  & 62.7& 65.6 &74.6 & 78.5 & 93.6 &72.5 &95.4& 86.2 & 71.9&  96.0& 78.0 &90.3 &80.7 &69.7 &76.8 &  80.6\\
SeENet  \cite{Pang2019TowardsBS}&98.7& 87.3& 93.7& 57.1& 61.8& 70.5& 77.6& 80.9& 94.0& 73.5& 95.9& 87.5& 71.6& 96.3& 76.4& 88.0& 79.9& 73.0&78.5&  81.2\\
ANL\cite{Zhu2019AsymmetricNN}& -& -& -& -& -& -& -& -& -& -& -& -& -& -& -& -& -& -& -&  81.3\\
CCNet  \cite{huang2019ccnet}& -& -& -& -& -& -& -& -& -& -& -& -& -& -& -& -& -& -& -&  81.4\\
BFP  \cite{Ding2019BoundaryAwareFP}&98.7& 87.0& 93.5& 59.8& 63.4& 68.9& 76.8& 80.9& 93.7& 72.8& 95.5& 87.0& 72.1& 96.0& 77.6& 89.0& 86.9& 69.2& 77.6&  81.4\\
DANet \cite{fu2019dual}&98.6 &86.1 & 93.5  &56.1 &63.3  &69.7& 77.3& 81.3& 93.9 &72.9& 95.7 &87.3&  72.9& 96.2& 76.8&  89.4& 86.5& 72.2 &78.2 &81.5\\
GANet  \cite{Zhang2019DeepGA}& -& -& -& -& -& -& -& -& -& -& -& -& -& -& -& -& -& -& -&  81.6\\
SPYGR  \cite{Li2020SpatialPB}& 98.7& 86.9& 93.6& 57.6& 62.8& 70.3& 78.7& 81.7& 93.8& 72.4& 95.6& 88.1& 74.5& 96.2& 73.6& 88.8& 86.3& 72.1& 79.2&  81.6\\
OCNet  \cite{Yuan2018OCNetOC}& -& -& -& -& -& -& -& -& -& -& -& -& -& -& -& -& -& -& -&  81.7\\
ACFNet\cite{zhang2019acfnet}& 98.7& 87.1& 93.9& 60.2& 63.9& 71.1& 78.6& 81.5& 94.0& 72.9& 95.9& 88.1& 74.1& 96.5& 76.6& 89.3& 81.5& 72.1& 79.2&  81.8\\
OCR \cite{Yuan2020ObjectContextualRF}& -& -& -& -& -& -& -& -& -& -& -& -& -& -& -& -& -& -& -&  81.8\\
SPNet  \cite{hou2020strip}& -& -& -& -& -& -& -& -& -& -& -& -& -& -& -& -& -& -& -&  82.0\\
DGCNet  \cite{Zhang2019DualGC}& 98.7& 87.4& 93.9& 62.4& 63.4& 70.8& 78.7& 81.3& 94.0& 73.3& 95.8& 87.8& 73.7& 96.4& 76.0& 91.6& 81.6& 71.5& 78.2&  82.0\\
HANet \cite{Choi2020CarsCF}& \textbf{98.8}& 88.0& 93.9& 60.5& 63.3& 71.3& 78.1& 81.3& 94.0&72.9&\textbf{96.1}& 87.9& 74.5& 96.5& 77.0& 88.0& 85.9& 72.7& 79.0&  82.1\\
RecoNet  \cite{Chen2020TensorLR}& -& -& -& -& -& -& -& -& -& -& -& -& -& -& -& -& -& -& -&  82.3\\
ACNet \cite{Fu2019AdaptiveCN}&98.7 & 87.1&93.9&61.6& 61.8& 71.4& 78.7& 81.7& 94.0& 73.3& 96.0& 88.5& 74.9& 96.5& 77.1& 89.0& 89.2 &71.4& 79.0& 82.3\\
GFF \cite{Li2020GatedFF}&98.7& 87.2& 93.9 &59.6& 64.3 &71.5 &78.3 &82.2& 94.0 &72.6 & 95.9 & 88.2& 73.9 & 96.5 &79.8 &92.2 &84.7& 71.5  & 78.8&  82.3\\
OCR \cite{Yuan2020ObjectContextualRF}& -& -& -& -& -& -& -& -& -& -& -& -& -& -& -& -& -& -& -&  82.4\\
DRANet \cite{Fu2021SceneSW}& \textbf{98.8}& 87.6& 94.1& 61.7& 62.7& \textbf{72.9}& \textbf{80.0}& \textbf{83.0}&\textbf{94.2}& 73.8& 96.0& \textbf{88.8}&\textbf{76.1}& \textbf{96.6}& 76.6& 89.8& 88.0& \textbf{73.8}&\textbf{80.0}&  82.9\\
\midrule[0.5pt]
\textbf{Denoised NL}&\textbf{98.8} & \textbf{87.9}  & \textbf{94.2}   & \textbf{64.1} &\textbf{65.0} & 72.3 & 78.7 & 82.7 &\textbf{94.2} &  \textbf{73.9}  & \textbf{96.1} &88.2   & 75.1  & 96.5  &\textbf{81.4} &\textbf{94.1}& \textbf{90.8} & 73.6  &79.3  & \textbf{83.5} \\
\bottomrule[1pt]
\end{tabular}}
\end{center}
\label{Citys-SOTA}
\end{table*}

\begin{figure*}[htp]
	\center
	\includegraphics[width=1\textwidth]{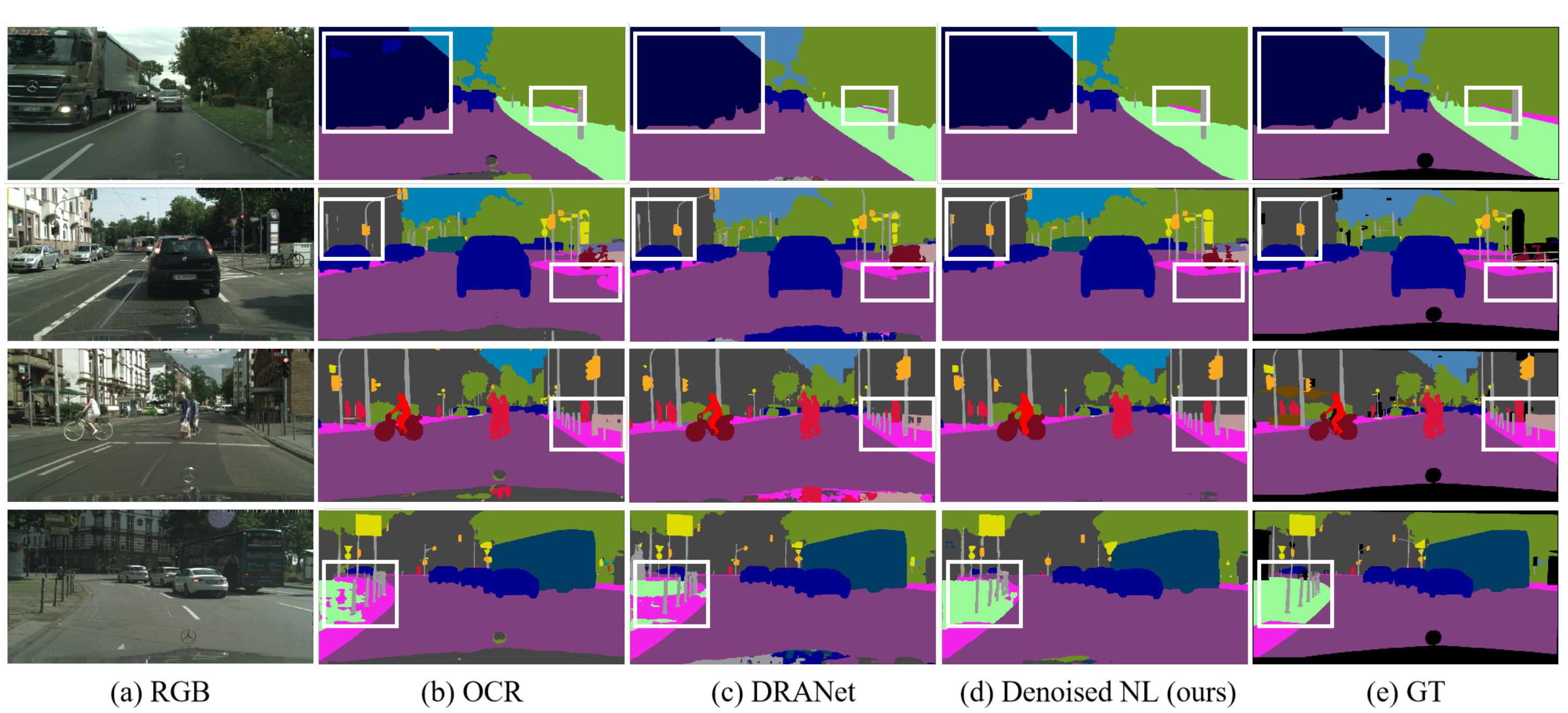}
	\caption[stereo]{Qualitative comparisons with the state-of-the-art models on the Cityscapes val set. Particularly, we use the official code and the provided model weights of these two methods for visualization. From left to right: (a) RGB inputs, (b) results obtained by OCR \cite{Yuan2020ObjectContextualRF}, (c) results generated by DRANet \cite{Fu2021SceneSW}, (d) the proposed Denoised NL, and (e) Ground Truth (GT). White squares denote the challenging regions.
}
	\label{fig-qualitive1}
\end{figure*}

\subsection{Implementation Details}

Our implementation is based on PyTorch \cite{Paszke2017AutomaticDI}, and we train the network using standard Stochastic Gradient Descent~(SGD)~\cite{krizhevsky2012imagenet} with the momentum of 0.9. Following the prior work~\cite{Yu2018LearningAD}, we apply the poly learning rate policy in which the initial learning rate is decayed by \((1-\frac{iter}{max\_iter})^{power}\) with power=0.9. Besides, random color jittering, random horizontal flipping, random cropping, and random scaling with the range of [0.5, 2] are adopted for data augmentation. The data augmentation scheme makes our network resist overfitting. Moreover, we use the synchronized batch normalization in all experiments for better mean/variance estimation due to the limited number of images hosted in each GPU. 

\begin{itemize}
\item \textit{Cityscapes} \cite{cordts2016cityscapes}: We set the initial learning rate as \(1e^{(-2)}\), weight decay coefficient as \(5e^{(-4)}\) , crop size as~\(768\times 768\), mini-batch size as 8. For the experiments evaluated on val/test set, we set training epochs as 240/480 on train/train+val set. 
\item \textit{ADE20K} \cite{Zhou_2017_CVPR}: We set the initial learning rate as \(2e^{(-2)}\), weight decay coefficient as \(1e^{(-4)}\), crop size as \(608\times608\), mini-batch size as 16 and and training epochs as 240.
\end{itemize}

\subsection{Experiments on the Cityscapes Dataset}

\textbf{Dataset and Evaluation Metrics.}  Cityscapes \cite{cordts2016cityscapes} is a dataset for urban scene segmentation. It contains 5000 images with fine pixel-level annotations from 50 cities in different seasons. The 5000 images with fine annotations are further divided into sets with 2975, 500, and 1525 images for training, validation, and testing, respectively. Cityscapes defines 19 categories containing both stuff, and objects for semantic segmentation and each image is with \(1024\times2048\) resolution. Also, 20,000 coarsely annotated images are provided in the dataset. It is noted that we only used fine data for training in our work. For evaluation, we adopt the mean of class-wise Intersection over Union (mIoU).

\begin{table}[t]
\caption{Ablation studies on the Cityscapes val set based on the ResNet-50 (R-50) and ResNet-101 (R-101).}
\begin{center}
\normalsize
\begin{tabular}{lcccc}
\toprule[2pt]
Method & Backbone & GR &  LR &mIoU (\%)\\
\midrule[2pt]
 Baseline & R-50 & &&76.7\\
 Denoised NL  & R-50 &\(\surd\) & & 81.2\\
Denoised NL  &R-50&   &\(\surd\)&81.4\\
Denoised NL&R-50 & \(\surd\)&\(\surd\)&82.0\\
\midrule[1pt]
 Baseline & R-101 & &&77.9\\
 Denoised NL  & R-101 &\(\surd\) &  &82.3\\
Denoised NL  &R-101&   &\(\surd\)&82.5\\
Denoised NL&R-101 & \(\surd\)&\(\surd\)&83.5\\
\bottomrule[1pt]
\end{tabular}
\end{center}
\label{table2}
\end{table}

\begin{figure}[t]
	\center
	\includegraphics[width=1\columnwidth]{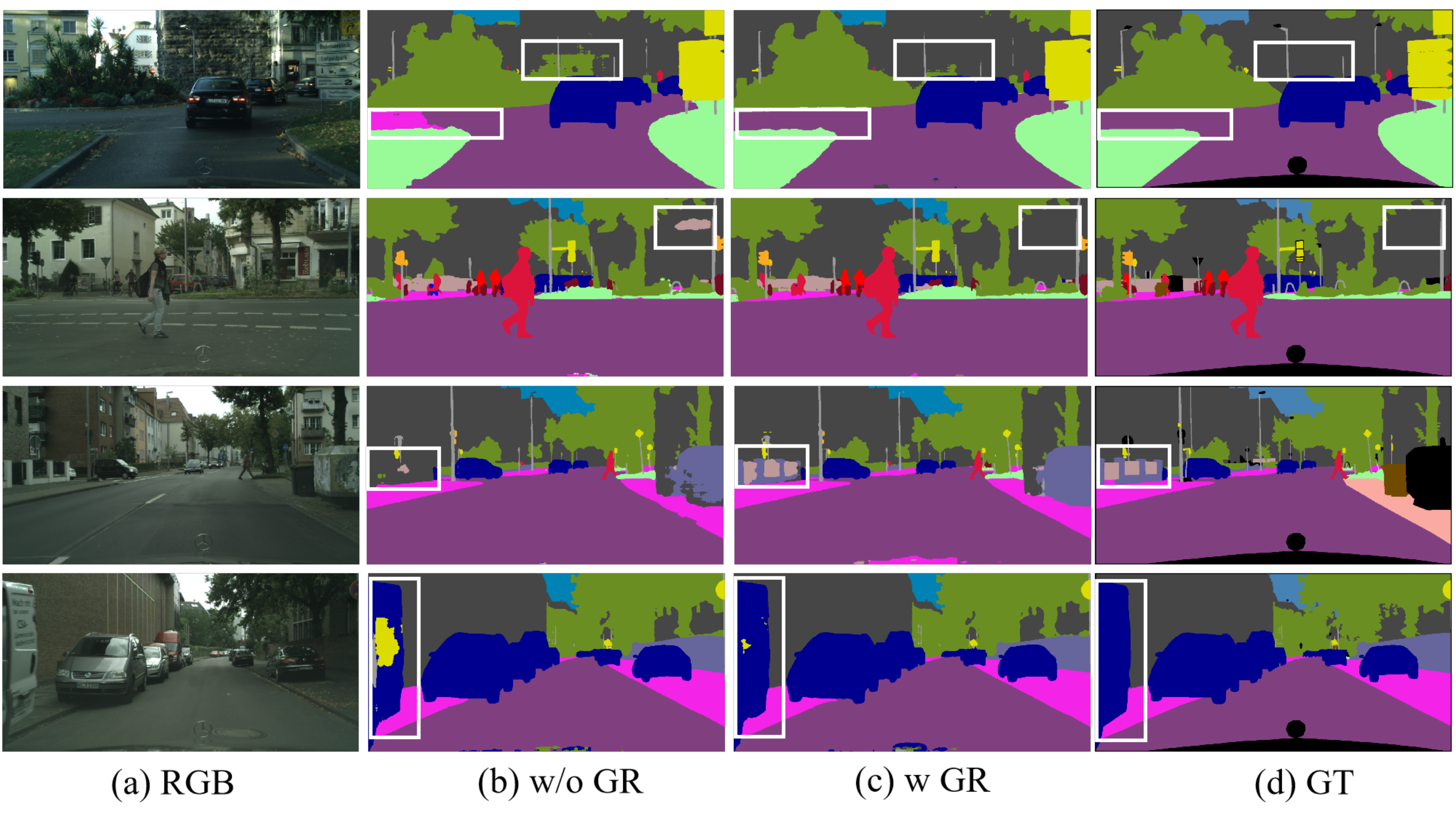}
	\caption[stereo]{Visualization results of Global Rectifying block on Cityscapes val set.
}
	\label{fig-ablation1}
\end{figure}

\begin{figure}[htp]
	\center
	\includegraphics[width=1\columnwidth]{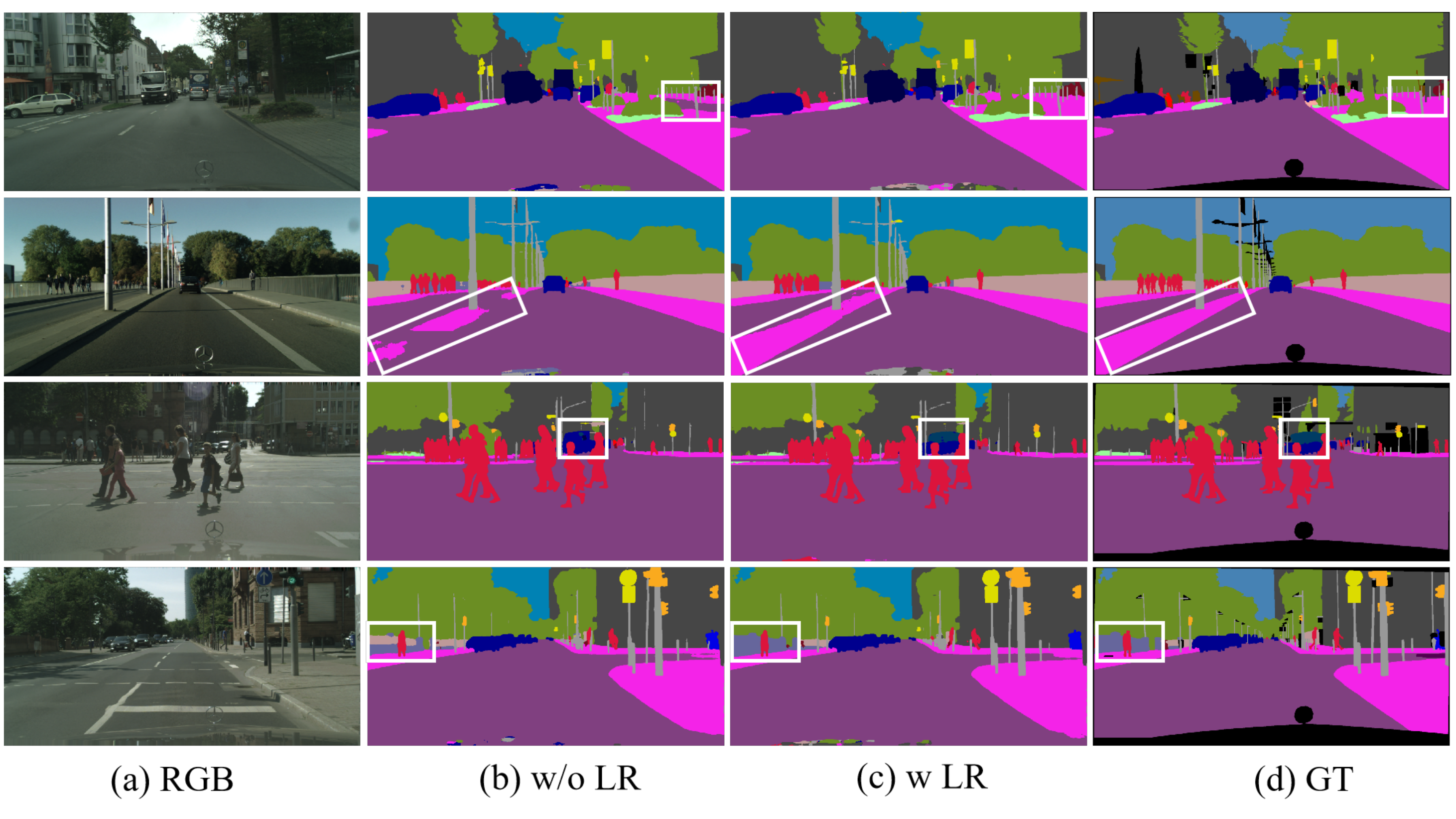}
	\caption[stereo]{Visualization results of Local Retention block on Cityscapes val set.
}
	\label{fig-ablation2}
\end{figure}

\textbf{Comparison with State-of-the-Art Models. } 
The comparisons between state-of-the-art methods and our proposed Denoised NL on the Cityscapes dataset are summarized in Table \ref{Citys-SOTA}. For a fair comparison, we only present state-of-the-art results without using extra coarse annotated data. In this paper, we set the most common-utilized ResNet-101 as our backbone network, and we can observe that our method outperforms existing approaches and can significantly achieve the state-of-the-art result of \textbf{83.5}\% even with the smallest resolution (\textbf{16$\times$}). Besides, from the detailed per-category comparison results, we can see that our method achieves the highest IoU on \textbf{11} out of 19 categories, and significant improvements are made on both small/thin and large/thick categories such as \emph{fence}, \emph{bus}, and \emph{train}. Compared with the non-local based methods such as CCNet, DANet, and DRANet, our method achieves better segmentation accuracy, which can also prove the effectiveness of Denoised NL in filtering the attention noises.

Furthermore, to demonstrate the advantages of Denoised NL, some qualitative comparisons are carried out against the two state-of-the-art methods (i.e., OCR and DRANet) on the Cityscapes validation set, as shown in Fig. \ref{fig-qualitive1}. It is obvious that our proposed Denoised NL can improve the feature discriminative ability and also fill in the hollows with different sizes or various shapes. For example, no matter it is a large object \emph{truck} (row 1) or a slender \emph{pole} (row 2 left square), our model can handle it well. The reason may lie in that our method can not only exploit the class-level predictions to distinguish the inter-class and intra-class context but also capture pixel-neighbored dependencies to obtain a more focused view and thus maintain the segmentation consistency and smoothness. 

\begin{table}
\caption{Comparison with multi-scale context modeling models on the Cityscapes val set. }
\begin{center}
\normalsize
\begin{tabular}{l|c}
\toprule[2pt]
Method & mIoU (\%)\\
\midrule[2pt]
Baseline (R-50) & 76.7\\
+PPM  & 77.6 \\
+ASPP   & 77.9  \\
\textbf{+Denoised NL} & \textbf{82.0} \\
\midrule[0.5pt]
Baseline (R-101) &77.9 \\
+PPM  & 79.0 \\
+ASPP  & 79.2  \\
\textbf{+Denoised NL} & \textbf{83.5} \\
\bottomrule[1pt]
\end{tabular}
\end{center}
\label{table4}
\end{table}

\begin{figure*}[htp]
	\center
	\includegraphics[width=1\textwidth]{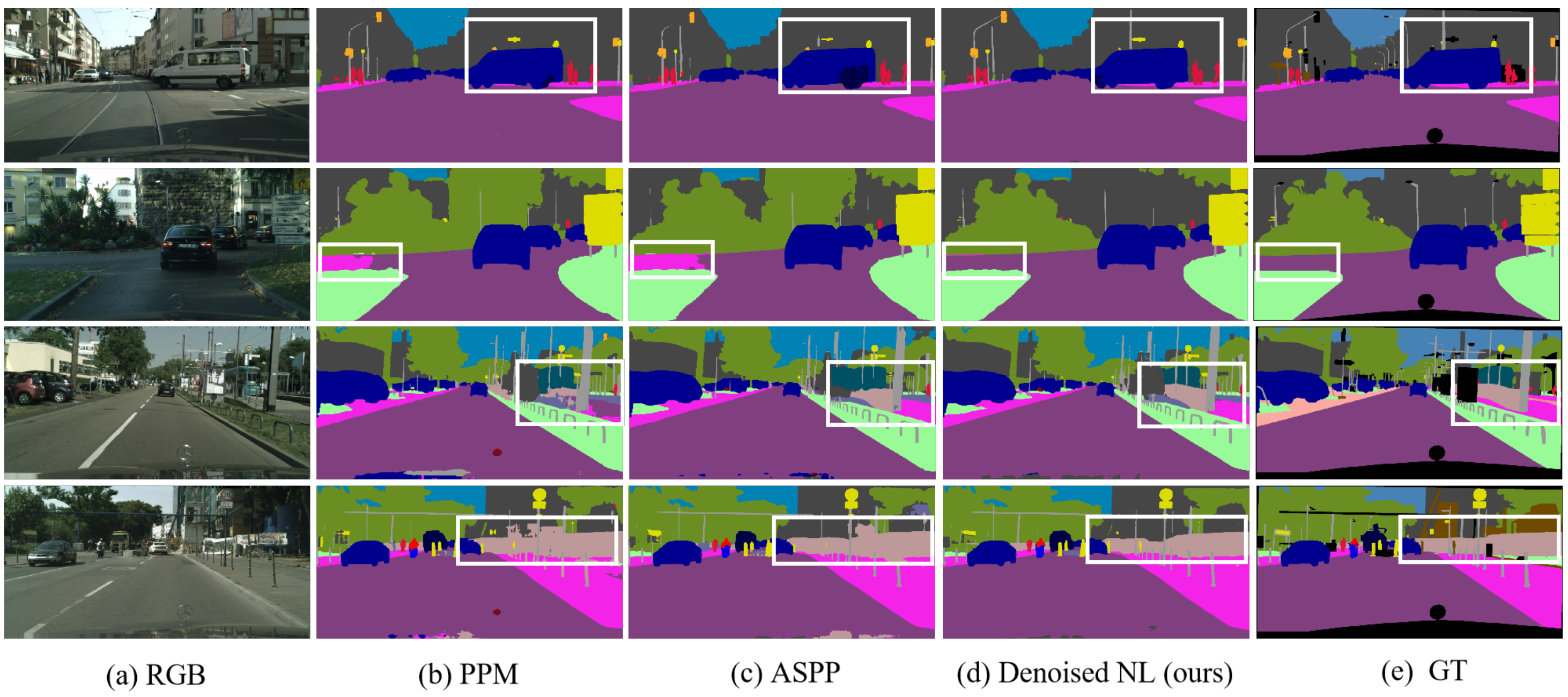}
	\caption[stereo]{Qualitative comparisons with multi-scale context modeling modules on the Cityscapes val set. From left to right: (a) RGB inputs, (b) results obtained by PPM \cite{zhao2017pyramid}, (c) results generated by ASPP \cite{Yuan2020ObjectContextualRF}, (d) our Denoised NL, and (e) ground truth. White squares denote the challenging regions.
}
	\label{fig-multi_scale}
\end{figure*}

\begin{table}
\caption{Comparisons with existing non-local based models on Cityscapes val set. `*' denotes that we only use CPAM and CCAM module of DRANet in our re-implementation, because we only compare with the non-local based modules.}
\begin{center}
\normalsize
\begin{tabular}{l|c}
\toprule[2pt]
Method & mIoU (\%) \\
\midrule[2pt]
Baseline (R-50) & 76.7 \\
+NL&  79.5\\  
+OCNet &80.3 \\
+CCNet &79.9 \\  
+DANet  & 80.3 \\  
+DRANet*  &  80.2 \\  
\textbf{+Denoised NL} & \textbf{82.0}\\
\midrule[0.5pt]
Baseline (R-101) & 77.9 \\
+NL& 80.6  \\  
+OCNet & 81.3  \\
+CCNet  &81.1 \\  
+DANet  &  81.2 \\  
+DRANet*  &  81.1 \\  
\textbf{+Denoised NL} &\textbf{83.5}\\
\bottomrule[1pt]
\end{tabular}
\end{center}
\label{table-nl}
\end{table}

\textbf{Ablation Study}
%
We conduct the ablation study on the Cityscapes validation set to verify the effectiveness of each component of our method. And the ablation studies are summarized in Table \ref{table2}. We use ResNet-50 and ResNet-101 as our backbone network and the Baseline is set to learn without any attention strategy. As presented in the upper part of Table \ref{table2}, the Baseline just gets mIoU of 76.7\% on semantic segmentation with ResNet-50. Applying GR block can obtain 4.5\% mIoU gain compared with the Baseline, demonstrating the effectiveness of eliminating the inter-class noises of GR. Also, if we only add LR, we can get 4.7\% accuracy improvement, which can convincingly confirm the significance of LR. Finally, with both GR and LR block, our proposed Denoised NL can significantly achieve 82.0\% by a large margin (5.3\%), and this result proves GR and LR can bring mutual benefits to the semantic segmentation task. Furthermore, when we employ ResNet-101 as the backbone, our network can also significantly improve the segmentation performance over the Baseline by 5.6\%.

Moreover, we provide some visualization examples of our ablation studies in Fig. \ref{fig-ablation1} and Fig. \ref{fig-ablation2}. It can be observed that with GR our model can correctly distinguish the misclassified category and clearly identify some details and object boundaries. Besides, after the usage of LR, the inconsistency inside large objects can be better handled. 

\begin{figure*}[htp]
\centering
\includegraphics[width=1\textwidth]{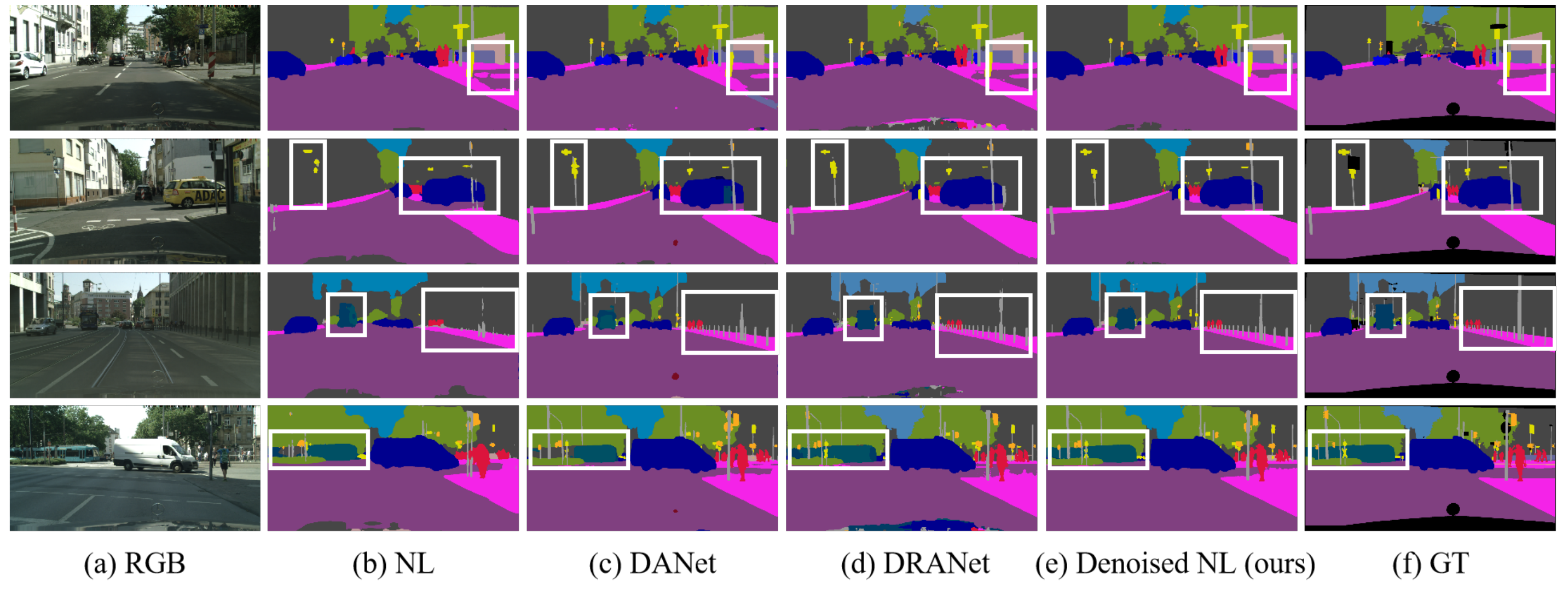}
\caption{Qualitative comparisons against different self-attention modules on the Cityscapes val set. We replace the proposed Denoised NL with the corresponding attention modules of other works and then visualize the final results of our re-implementations.}
\label{fig-self-attention}
\end{figure*}

\begin{figure*}[htp]
\centering
\includegraphics[width=1\textwidth]{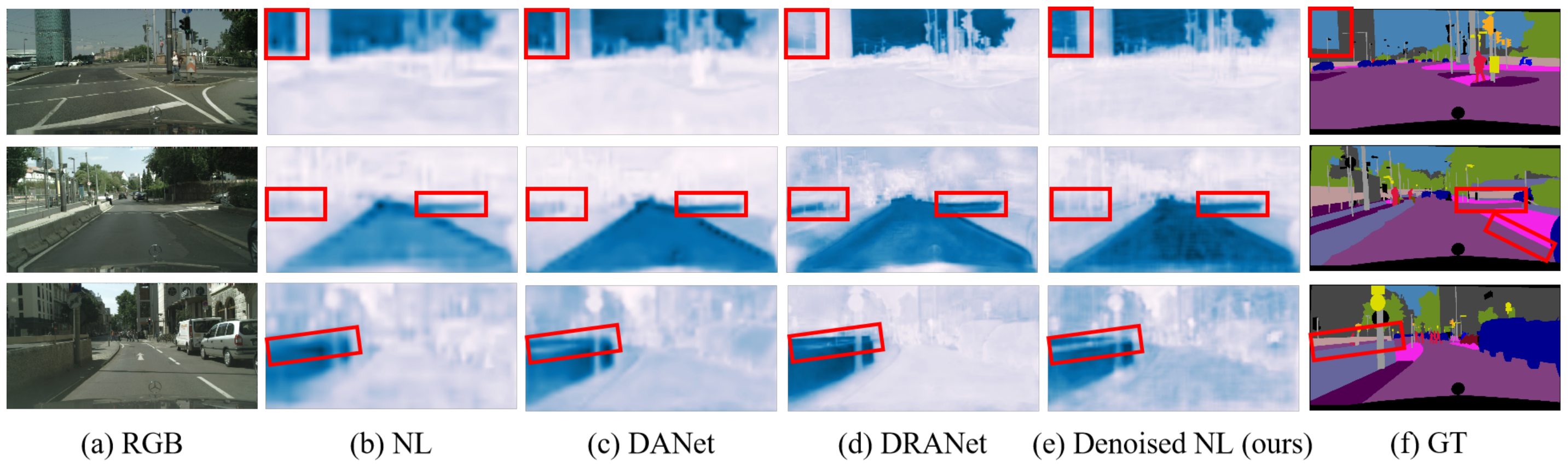}
\caption{Visualization of the predicted regions on the Cityscapes val set.}
\label{fig-regions}
\end{figure*}

\begin{table}
\caption{Complexity comparison. All the numbers are estimated for an input of \(1\times 3\times 768\times 768\). `*' denotes that we only use CPAM and CCAM module of DRANet in our re-implementation, because we only compare with the non-local based modules.}
\begin{center}
\normalsize
\resizebox{\columnwidth}{!}{
\begin{tabular}{l|ccc}
\toprule[2pt]
Method & GFLOPs ($\Delta$)&Memory ($\Delta$) &mIoU (\%)\\
\midrule[2pt]
Baseline & - &- &77.9  \\
+NL& 7.63G & 407M& 80.6 \\  
+CCNet& 1.77G & 52M& 81.1 \\  
+DANet  & 10.04G &570M&  81.2\\  
+DRANet*  &\textbf{0.76G} &\textbf{30M}& 81.1 \\  
\textbf{+Denoised NL} & 7.83G  &428M& \textbf{83.5} \\
\bottomrule[1pt]
\end{tabular}}
\end{center}
\label{table3}
\end{table}

\textbf{Comparison with Multi-Scale Context Modeling Models. }
In this section, the comparisons with multi-scale context modeling modules (i.e., PPM \cite{zhao2017pyramid} and ASPP \cite{chen2017deeplab}) are introduced. 
Table \ref{table4} presents the quantitative results on the Cityscapes val set, and the corresponding visualization results are shown in Fig. \ref{fig-multi_scale}. From the comparison results, it can be seen that our Denoised NL outperforms these multi-scale context schemes by a large margin. Since we use features with stride \textbf{16} as the input of our proposed module, we can also find that when the spatial size is relatively small, the benefits of those multi-scale context modeling modules would be limited. In contrast, our approach can avoid the accuracy fluctuation and model both global and local contexts under different spatial sizes.

\textbf{Comparison with Existing Self-Attention Models. }
To further highlight the advantages of our proposed method, we compare Denoised NL with the existing non-local based models, as shown in Table \ref{table-nl}. Obviously, Denoised NL shows superior results on all existing non-local works with the highest accuracy of 83.5\%. For a fair comparison, we re-implement several well-known models with our pipeline by replacing the Denoised NL with the corresponding attention module. Note that in our re-implementation all these non-local methods use \textbf{16$\times$} features as the module input, so there might be some accuracy fluctuations. Although the re-implemented 16$\times$ attention models has a relatively lower performance than the released 8$\times$ model, we can observe that when the attention module is changed to our Denoised NL, the performance has been greatly improved, and it far exceeds all the existing self-attention methods. Besides, the corresponding qualitative comparisons against different attention modules are shown in Fig. \ref{fig-self-attention}, which also proves the superiority of our method.

To get a deeper understanding of our Denoised NL, we further visualize the predicted regions in Fig. \ref{fig-regions}, and we also report the results of NL, DANet, and DRANet for comparisons. It can be observed that the predicted regions of our approach are more accurate and intact. After applying our Denoised NL, both intra-class consistency and inter-class differentiability are increased. For example, the predicted regions with other methods usually exist hollows and unsmooth boundaries, such as the \textit{sky} in row 1, and the boundary between the \textit{road} and \textit{sidewalk} in row 2, while our method can avoid these mistakes, which might be the main advantages of our approach.

As for the computational complexity, we measure the number of FLOPs and GPU memory that are introduced by the non-local modules and do not count the complexity from the baseline network. The comparison results in Table \ref{table3} show that our Denoised NL can bring the biggest performance gain with the less computational cost.

\begin{figure*}[htp]
\centering
\includegraphics[width=1\textwidth]{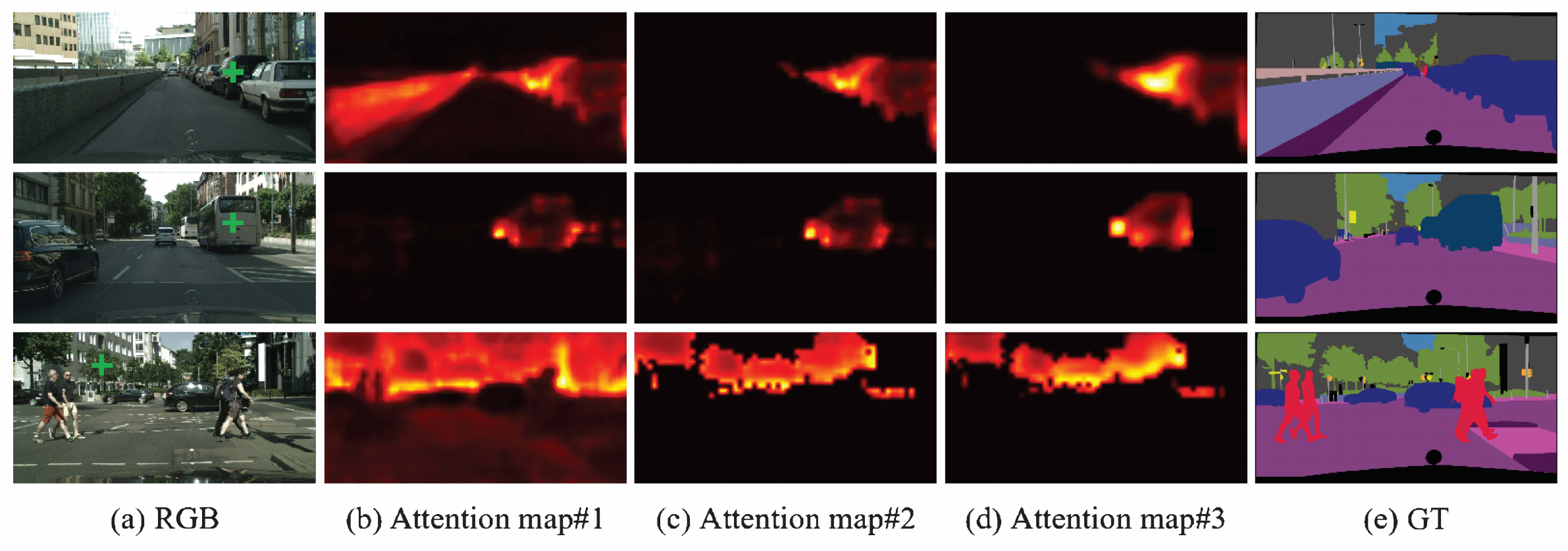}
\caption{Visualization of attention maps on the Cityscapes val set.}
\label{maps}
\end{figure*}

\textbf{Visualization of Attention Map. } In this subsection, we visualize the attention maps of our method in Fig. \ref{maps} to further explain how Denoised NL works. For each input image, we select one pixel (marked with a green cross) and show its corresponding attention maps. Specifically, there are three attention maps in our model, i.e., the original one, map after inter-class denoising, and map after intra-class denoising. We denote these three maps as attention map \#1, \#2, \#3 respectively, which correspond to \(A\), \(A'\), \(A''\) in Fig. \ref{DNL}. From the visualization results, we can find that the original attention map can capture long-range semantic dependencies, while with some inevitable inter-class and intra-class noises. After applying GR, the attention map becomes purer and only highlights the pixels belonging to the category of the selected pixel. Moreover, if we then add LR on the top of attention map~\#2, the attention hollows inside the object can be effectively filled. Take the second row in Fig. \ref{maps} as the example, after applying GR, the attention map changes from map \#1 to map~\#2, and the inter-class noises that do not belong to the class \textit{bus} are successfully filtered. Also, there are fewer attention hollows inside \textit{bus} in map \#3 after using LR.

\subsection{Experiments on the ADE20K Dataset}

\textbf{Dataset and Evaluation Metrics.} The ADE20K dataset~\cite{Zhou_2017_CVPR} is a challenging scene parsing benchmark. ADE20K is more challenging than other datasets with up to 150 stuff/object categories and diverse scenes with a total of 1,038 image-level labels. The challenge data is divided into 20K/2K images for training and validation. Also, images in this dataset are from different scenes with more scale variations. For evaluation, the mean of class-wise Intersection over Union (mIoU) is used.

\begin{table}[h]
\caption{Comparisons with state-of-the-art approaches on the ADE20K val set.}
\begin{center}
\normalsize
\begin{tabular}{lcc}
\toprule[2pt]
Method & Backbone & mIoU (\%)\\
\midrule[2pt]
RefineNet \cite{lin2017refinenet} & ResNet-101 &  40.20\\
PSPNet \cite{zhao2017pyramid} & ResNet-101 &  43.29\\
DSSPN \cite{liang2018dynamic} & ResNet-101 &   43.68\\
PSANet \cite{zhao2018psanet} & ResNet-101 &  43.77\\
SAC \cite{Zhang2017ScaleAdaptiveCF}& ResNet-101 &   44.30\\
EncNet \cite{Zhang2018ContextEF} & ResNet-101 &  44.65\\
DANet \cite{fu2019dual} & ResNet-101   &  45.22\\
CCNet \cite{huang2019ccnet} & ResNet-101   & 45.22\\
ANL \cite{Zhu2019AsymmetricNN} & ResNet-101   &  45.24\\
OCR \cite{Yuan2020ObjectContextualRF} & ResNet-101   &  45.28\\
GFF \cite{Li2020GatedFF}&ResNet-101&  45.33\\
DMNet \cite{He2019DynamicMF} & ResNet-101 &    45.50\\
RecoNet \cite{Chen2020TensorLR}& ResNet-101  &45.54\\
ACNet \cite{Fu2019AdaptiveCN}  & ResNet-101 &   45.90 \\
AlignSeg \cite{Huang2021AlignSegFS} & ResNet-101 &    45.95\\
DNL \cite{Yin2020DisentangledNN} & ResNet-101 &   45.97\\
DRANet \cite{Fu2021SceneSW}& ResNet-101 & 46.18\\
HRNet \cite{Sun2019DeepHR} & HRNetV2-W48  &  42.99\\
DNL  \cite{Yin2020DisentangledNN} & HRNetV2-W48& 45.82\\
\midrule[0.5pt]
\textbf{Denoised NL} & ResNet-101 & \textbf{46.69}\\
\bottomrule[1pt]
\end{tabular}
\end{center}
\label{table5}
\end{table}

\textbf{Results on ADE20K Dataset.} We conduct experiments on the ADE20K dataset to validate the effectiveness of our
method. The comparative results between our Denoised NL and the previous excellent models on the ADE20K val set are reported in Table \ref{table5}. We can see that our approach can achieve the state-of-the-art result with 46.69\% mIoU.

\section{Conclusions}

\label{5}
In this paper, we focus on filtering the attention noises contained in the traditional non-local methods. We first give a brief definition and classification for the attention noises. After that, we further theoretically analyze the causes of noise and the possible solutions. To this end, we propose a novel Denoised NL, including two primary components: Global Rectifying~(GR) block and Local Retention (LR) block to eliminate inter-class noises and intra-class noises respectively. And we are the first one to try to improve the traditional non-local mechanism from the view of attention map denoising. Specifically, GR utilizes class-level predictions to estimate a binary map, which can correctly identify whether two pixels belong to a category or not. With the use of GR, we can easily remove points that don't belong to the selected category, i.e., inter-class noises. Moreover, for the intra-class noises, we believe that they can be successfully filtered if the attention module has a more focused contextual view. Thus LR generates a set of local regions and estimates the dependencies between pixels and their local regions to help correctly identify the right category of inner points. We have conducted extensive experiments on several semantic segmentation benchmarks and experimental results show the success of our Denoised NL.


\bibliographystyle{IEEEtran}
\bibliography{ieeeconf}

\end{document}